\documentclass[a4paper, conference]{IEEEtran}
\IEEEoverridecommandlockouts
\usepackage{cite}

\usepackage{amsmath,amssymb,amsfonts}
\usepackage{algorithm}
\usepackage{algpseudocode}
\usepackage{footnote}
\makesavenoteenv{algorithm}
\usepackage{graphicx}
\usepackage{textcomp}
\usepackage{xcolor}
\usepackage{float}
\usepackage{array}
\usepackage{tabu}
\usepackage{multirow}
\usepackage{flafter}
\usepackage{geometry}
\usepackage{breqn}
\usepackage{tabulary}
\usepackage{etoolbox}
\usepackage{subcaption}
\usepackage{xspace}
\usepackage{tikz}
\usepackage{textcomp}
\usepackage{hyperref}
\usepackage{lipsum}

\makeatletter
\AfterEndEnvironment{algorithm}{\let\@algcomment\relax}
\AtEndEnvironment{algorithm}{\kern2pt\hrule\relax\vskip3pt\@algcomment}
\let\@algcomment\relax
\newcommand\algcomment[1]{\def\@algcomment{\footnotesize#1}}

\renewcommand\fs@ruled{\def\@fs@cfont{\bfseries}\let\@fs@capt\floatc@ruled
  \def\@fs@pre{\hrule height.8pt depth0pt \kern2pt}%
  \def\@fs@post{}%
  \def\@fs@mid{\kern2pt\hrule\kern2pt}%
  \let\@fs@iftopcapt\iftrue}
\makeatother
\renewcommand\IEEEkeywordsname{Keywords}
\def\BibTeX{{\rm B\kern-.05em{\sc i\kern-.025em b}\kern-.08em
    T\kern-.1667em\lower.7ex\hbox{E}\kern-.125emX}}

\newcommand\copyrighttext{%
  \footnotesize \textcopyright 2019 IEEE. This paper is under review in "22nd International Conference on Computer and Information Technology (ICCIT), 2019". Personal use of this material is permitted.
  Permission from IEEE must be obtained for all other uses, in any current or future
  media, including reprinting/republishing this material for advertising or promotional
  purposes, creating new collective works, for resale or redistribution to servers or
  lists, or reuse of any copyrighted component of this work in other works.
}
\newcommand\copyrightnotice{%
\begin{tikzpicture}[remember picture,overlay]
\node[anchor=south,yshift=10pt] at (current page.south) {\fbox{\parbox{\dimexpr\textwidth-\fboxsep-\fboxrule\relax}{\copyrighttext}}};
\end{tikzpicture}%
}
    
\begin{document}

\title{HishabNet: Detection, Localization and Calculation of Handwritten Bengali Mathematical Expressions\\}

\author{\IEEEauthorblockN{ Md Nafee Al Islam}
\IEEEauthorblockA{\textit{Department of Electrical and Electronic Engineering} \\
\textit{Ahsanullah University of Science and Technology}\\
\textit{seiumiut@gmail.com}\\
\\
}\\[-6.0ex]
\and
\IEEEauthorblockN{Siamul Karim Khan}
\IEEEauthorblockA{\textit{Department of Computer Science and Engineering} \\
\textit{Bangladesh University of Engineering and Technology}\\
\textit{siamulkarim@gmail.com}
\\
}\\[-6.0ex]

}

\maketitle
\copyrightnotice
\begin{abstract}
Recently, recognition of handwritten Bengali letters and digits have captured a lot of attention among the researchers of the AI community. In this work, we propose a Convolutional Neural Network (CNN) based object detection model which can recognize and evaluate handwritten Bengali mathematical expressions. This method is able to detect multiple Bengali digits and operators and locate their positions in the image. With that information, it is able to construct numbers from series of digits and perform mathematical operations on them. For the object detection task, the state-of-the-art YOLOv3 algorithm was utilized. For training and evaluating the model, we have engineered a new dataset ``Hishab'' which is the first Bengali handwritten digits dataset intended for object detection. The model achieved an overall validation mean average precision (mAP) of 98.6\%. Also, the classification accuracy of the feature extractor backbone CNN used in our model was tested on two publicly available Bengali handwritten digits datasets: NumtaDB and CMATERdb. The backbone CNN achieved a test set accuracy of 99.6252\% on NumtaDB and 99.0833\% on CMATERdb.
\newline
\end{abstract}

\section{Introduction}
The advances in the field of Artificial intelligence and Machine learning have revolutionized computer vision. Efficient Deep Learning techniques are proving to be very impactful for image classification, object detection and pattern recognition. Convolutional Neural Networks (CNNs) are used widely to extract patterns and features from images and have shown exceptional performance in the field of handwritten digit recognition. Bengali is the 7th most spoken language in the world which is used as the first language in Bangladesh and some part of the India. In the past decade, a lot of research have been performed to recognize handwritten Bengali characters and digits. Handwritten digit recognition systems are very helpful in Optical Character Recognition (OCR) applications such as license plate identification, postal code identification, bank cheque reading etc. 
Object detection involves locating instances of objects in images or videos. It involves two tasks: Object Localization where the network identifies where the object is, putting a bounding box around it, and Classification where the network predicts the class of each object present in the image.\par
In this work, we have developed a CNN-based object detection system which can detect and localize handwritten Bengali mathematical expressions and perform mathematical operations. The model was built using a CNN-based architecture YOLOv3 \cite{yolov3} which uses the YOLO \cite{yolo} algorithm for object detection.  \par
All of the previous works on recognition of Bengali digits dealt only with the classification task. None of the works focused on localization of digits and thus, they were able to identify only one digit at a time. They were not able to localize multiple digits and mark their positions in the image. Our system treats the recognition of Bengali digits as an object detection task. Thus, our system can detect multiple digits and operators in a single image, localize them and mark their positions in the image and is, therefore, able to construct numbers from series of digits using their positions, recognize mathematical operators and hence perform mathematical operations on those numbers using the operators. The model can perform multiple mathematical operations given in a single image. It is also able to work with brackets and decimal numbers. \par
Also, as the previous works on Bengali handwritten digits recognition have all worked with object classification, their datasets do not contain localization information and has no images with multiple objects in it. Thus, those datasets were not suitable for object detection. For that reason, we have engineered a new dataset ``Hishab" which contains 40000 images of Bengali digits and operators along with their bounding box annotations. \par
We also show that the CNN architecture based on Darknet53 utilized as the backbone feature extractor for our network achieves a test set accuracy of 99.6252\% on NumtaDB and 99.0833\% on CMATERdb.
In summary, we have made the following contributions:
\begin{itemize}
\item{We have designed a model which can detect and localize multiple Bengali digits and operators in an image. Using the positions of the digits and operators, the model is able to construct numbers and perform mathematical operations on those numbers. The model is also able to evaluate multiple mathematical expressions given to it in a single image.}
\item{We have created a new dataset ``Hishab", intended for object detection, which contains 40000 images with their bounding box annotations in separate text files.}
\item{We have showed that the CNN architecture used as the backbone for our model outperforms previously suggested architectures for handwritten Bengali digit recognition.}
\end{itemize}

\section{Background and Related Works}

In the first part of this section, we shall discuss about previous works on the domain of handwritten digit recognition. Then, we move forward to the previous works on Bengali handwritten digits recognition. After that, we discuss about the previous works on mathematical expression recognition. Next, we explain the advantages our method has, compared to the previous works on Bengali digit recognition. Finally, we provide a concise overview of mean Average Precision (mAP), the metric that is typically used for evaluating object detection systems.

\subsection{Existing Works on Handwritten Digit Recognition}
Abdeljalil and Youcef used sliding windows detector for segmentation and recognition of connected digits in \cite{gattal2017segmentation}. They tested their method on NIST SD19 database. Liu et al. \cite{liu2003handwritten} tested seven classifiers each with eight different feature vectors on three popular datasets - CENPARMI, CEDAR, and MNIST. Out of the seven classifiers, the SVM classifier with RBF Kernel was the best in terms of accuracy. Ignat and Aciobanitei in \cite{ignat2016handwritten} used rotation and edge filtering to extract features from handwritten digits. They used k-NN and SVM classifier to test their method on the MNIST database and achieved an accuracy of 99\%. Kussul and Baidyk proposed a neural classifier for handwritten digit recognition based on permutative coding technique in \cite{kussul2003permutative}. The coding technique came from associative-projective neural networks. Their classifier was tested on MNIST database and it showed an error of 0.54\%. Ghosh and Maghari \cite{ghosh2017comparative} have compared the performance of Deep Neural Networks (DNN), Deep Belief Networks (DBN) and Convolutional Neural Networks (CNN) on handwritten digit detection and they found DNN to be the most accurate one with 98.08\% accuracy. Shamim et al. used different machine learning techniques such as Multilayer Perceptron, Support Vector Machine, Naive Bayes, Bayes Net, Random Forest, J48 and Random Tree for handwritten digit recognition in \cite{shamim2018handwritten}. Several works have been done to recognize handwritten characters and digits belonging to different languages such as Chinese \cite{yin2013icdar}, Arabic \cite{amin1998off}, Greek \cite{kavallieratou2001gruhd}, Japanese \cite{zhu2010robust} etc.

\subsection{Works on Bengali Handwritten Digit Recognition}
Obaidullah et al. proposed the HNSI (Handwritten Numeral Script Identiﬁcation) framework to detect handwritten scripts belonging to four different languages - Bengali, Roman, Urdu, and Devanagari \cite{obaidullah2015numeral}. Boni et al. proposed a methodology called Chemical Reaction Optimization (CRO) to detect handwritten Bengali digits in \cite{boni2018handwritten} and they achieved an accuracy of 98.96\% on CMATERdb. Rabby et al. proposed a lightweight CNN model called BornoNet to detect Bengali handwritten characters \cite{rabby2018bornonet}. They used three datasets for training and validation – BanglaLekha Isolated, CMATERdb and ISI. They achieved accuracy of 98\%, 96.81\% and 95.71\% on these three datasets respectively.

\subsection{Works on Mathematical Expression Recognition}
Shinde et al. proposed a neural network based algorithm to detect mathematical equations \cite{shinde2017new}. They used a feed forward back propagation neural network with gradient decent to construct their model. Garain and Chaudhuri proposed a method to understand online handwritten mathematical expressions \cite{garain2004recognition}.They used a combination to two different classifiers to improve the detection accuracy. Also they designed a context-free grammar to convert the input expressions to TEX strings and then to MathML format. Genoe and Kechadi developed a Real-time Recognition System for Handwritten Mathematics in \cite{genoe2010real}. They used a structural development technique where they used a grammar based approach to enable instantaneous user feedback. D’souza and Mascarenhas recognized offline handwritten mathematical expression using a Convolutional Neural Network \cite{d2018offline}.

\subsection{Comparison of our method with preceding works}

The existing works on Bengali Digit recognition used different machine learning techniques such as Support Vector Machines (SVM), K-nearest neighbors (KNN), Artificial Neural Networks (ANN), and Convolutional Neural Networks (CNN) to recognize and classify Bengali Handwritten digits. Some of them used morphological and graph-based approach for the task. None of the previous works on Bengali handwritten digits recognition carried out localization of digits and were only able to classify a single digit in an image. Our proposed method is able to detect and localize multiple digits and operators in an image which, in turn, allows us to detect and evaluate multiple mathematical expressions in an image.

\subsection{Evaluation Metric for Object Detection: mean Average Precision (mAP)}
\begin{figure}[ht]
    \centering 
    \includegraphics[width=0.9\linewidth]{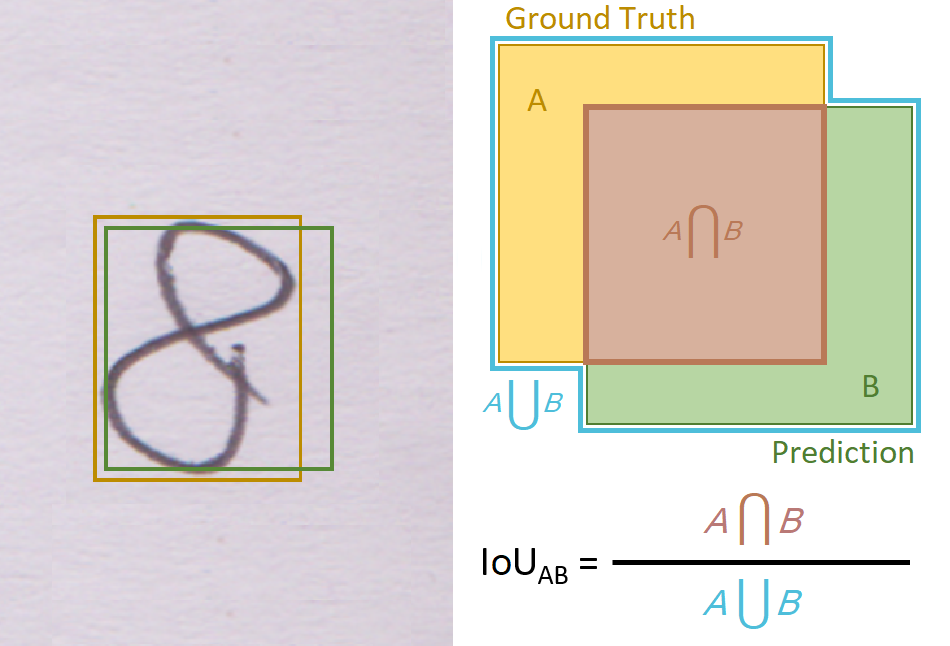}
    \caption{Calculation of Intersection over Union}
    \label{IOUexplain}
\end{figure}

In object detection, evaluation is non-trivial, because there are two discrete aspects that we need to consider:
\begin{itemize}
    \item \textbf{Localization:} Evaluating the location coordinates obtained from the model, for the object in the image.
    \item \textbf{Classification:} Ascertaining whether the object in the image is assigned to the correct class.
\end{itemize}
In order to evaluate the model on the task of object localization, we have to determine how accurately the model predicted the location of the object. This evaluation is done using the Intersection over Union threshold (IoU). It is the ratio of intersection to union of two boxes: the predicted bounding box and the ground-truth bounding box. Fig. \ref{IOUexplain} provides an example of IoU calculation. 
\par
Additionally, in a typical dataset there will be a lot of classes and the distribution of these classes might not be uniform. So, a simple accuracy-based metric can introduce biases in the evaluation. It is also essential to determine the likelihood of misclassifications. Thus, it is imperative to assign a “confidence score” with each detected bounding box and to analyze the model at various levels of confidence. To satisfy these needs, the Average Precision (AP) metric was introduced. 
Initially, we need to find the number of True Positives (TP), False Positives (FP) and False Negatives (FN). Assume that the IoU threshold is $t$. Table \ref{calMetrics} shows how the above quantities are computed. \par
\begin{table}[ht]
\bgroup
\centering
\def\arraystretch{1.1}
\setlength\tabcolsep{20pt}
\begin{tabular}{|c|c|c|}
\hline
IoU Condition                & Classification & Category \\ \hline
$\mbox{IoU}>t$     & Correct        & TP       \\ \hline
$\mbox{IoU}\leq t$ & Correct        & FP       \\ \hline
$\mbox{IoU}>t$     & Incorrect      & FN       \\ \hline
\end{tabular}
\egroup
\caption{Calculation of TP, FP and FN for object detection.}
\label{calMetrics}
\end{table}
These values are used to calculate precision ($\frac{TP}{TP+FP}$) and recall ($\frac{TP}{TP+FN}$) for the model. By varying the model's bounding box confidence score threshold, a precision-recall curve is constructed from the model’s detection outputs. The AP score is the area under this curve and is usually calculated as the mean precision at the set of 11 equally-spaced recall values, $Recall_i = [0, 0.1, 0.2, ..., 1.0]$. Thus,
\begin{equation}
AP = \frac{1}{11}\sum_{Recall_{i}} Precision\mbox{ }at\mbox{ }Recall_{i}
\end{equation}
The $Precision$ $at$ $Recall_{i}$ is taken to be the maximum precision measured at a recall exceeding $Recall_i$. The mean Average Precision or mAP score is calculated by averaging the AP score for all classes and/or for all IoU thresholds.

\section{Our Proposed Methodology}

Our work consists of two steps. First, we employ our object detection model based on YOLOv3 to localize and classify Bengali digits and mathematical operators in an image. Then, we group together these digits and operators to identify and evaluate mathematical expressions. Our system has been implemented using Darknet \cite{darknet13} an open source neural network framework.

\subsection{Object Detection and the YOLO Model}

Recently, CNN-based object detection algorithms have shown remarkable accuracy and speed. In the recent past, researchers have developed several efficient CNN-based object detection algorithms such as R-CNN \cite{girshick2014rich}, Fast R-CNN \cite{Girshick_2015_ICCV}, Faster R-CNN \cite{NIPS2015_5638}, SSD \cite{liu2016ssd}, RetinaNet \cite{lin2017focal} and YOLO \cite{yolo}. Before YOLO all the object detection models carried out detection in two stages. First, a set of regions of interests is proposed by selective search or regional proposal network. Then, these region candidates are processed by a classifier. But YOLO framed this as a regression problem and performed detection as well as classification using a single neural network.\par
Even though YOLO, being a one-stage detector, was the fastest among the other object detection algorithms, it lacked in accuracy and did a lot of localization errors, especially for small objects.  Later, Redmon et al. tweaked the architecture, using batch normalization, anchor boxes, multi-scale training, residual connections and many other techniques, to improve accuracy in YOLOv2 \cite{yolov2} and then in YOLOv3 \cite{yolov3}.  YOLOv3 has accuracy on par with the contemporary object detection algorithms while being faster than the other algorithms by a wide margin. So, we have employed the YOLOv3 architecture to carry out our task. Table \ref{compObDet} summarizes the accuracy and speed comparison between the different object detection algorithms.\par
\begin{figure}
    \centering
    \includegraphics[height=14cm, width=0.8\linewidth]{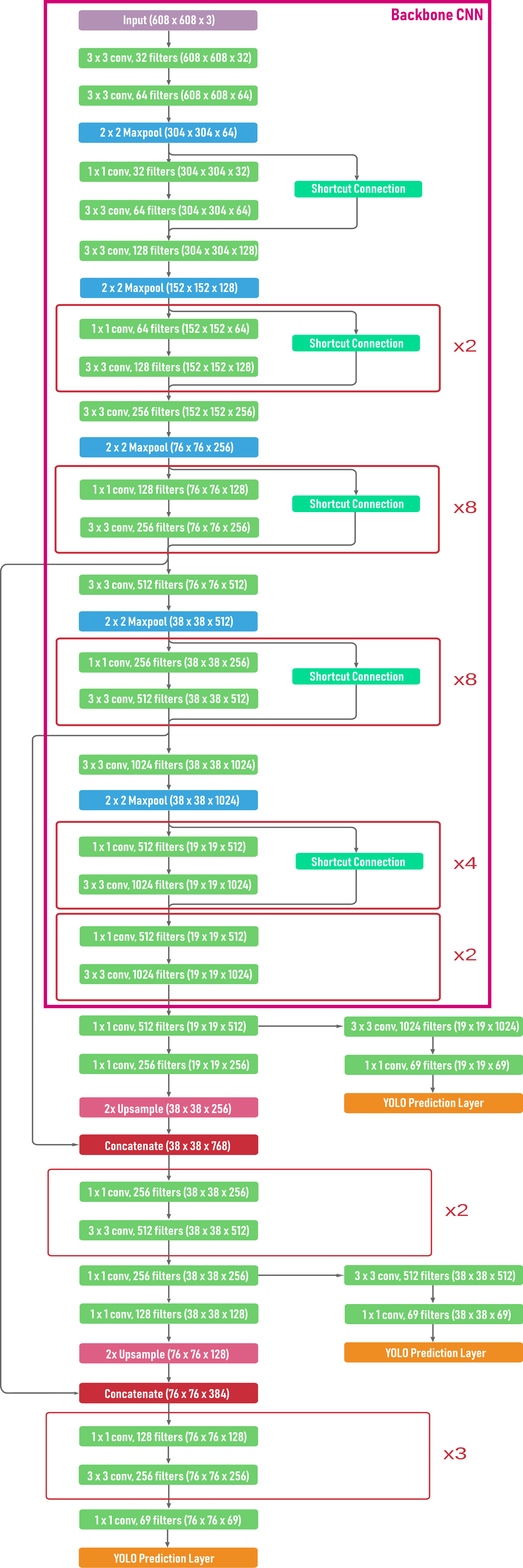}
    \caption{The overall architecture of the proposed YOLOv3 model.}
    \label{proposedmodel}
\end{figure}

\begin{table}[ht]
\bgroup
\centering
\def\arraystretch{1.1}
\setlength\tabcolsep{9pt}
\begin{tabular}{|c|c|c|}
\hline
Method            & \begin{tabular}[c]{@{}l@{}}COCO mAP, IoU=50\% (\%)\end{tabular} & time(ms) \\ \hline
SSD321            & 45.4                                                                  & 61       \\ \hline
DSSD321           & 46.1                                                                  & 85       \\ \hline
R-FCN             & 51.9                                                                  & 85       \\ \hline
SSD513            & 50.4                                                                  & 125      \\ \hline
DSSD513           & 53.3                                                                  & 156      \\ \hline
RetinaNet-50-500  & 50.9                                                                  & 73       \\ \hline
RetinaNet-101-500 & 53.1                                                                  & 90       \\ \hline
RetinaNet-101-800 & 57.5                                                                  & 198      \\ \hline
YOLOv3-320        & 51.5                                                                  & 22       \\ \hline
YOLOv3-416        & 55.3                                                                  & 29       \\ \hline
YOLOv3-608        & 57.9                                                                  & 51       \\ \hline
\end{tabular}
\egroup
\vspace{5pt}
\caption{A comparison of different object detection algorithms. Inference times are measured on a Pascal Titan X GPU.}
\label{compObDet}
\end{table}
The YOLO model divides the input image into an $S\times S$ grid. The value of $S$ depends on the scale at which the model is predicting the bounding box.  Each grid cell is responsible for detecting all objects whose center falls within its boundaries. It predicts 3 bounding boxes and confidence scores for those boxes. Each bounding box consists of five predictions: $x$, $y$, $w$, $h$, and confidence. The $(x, y)$ coordinates represent the center of the bounding box relative to the boundaries of the grid cell. The $w$ and $h$ variables represent the width and height of the object normalized according to the width and height of the whole image. Finally the confidence prediction represents the Intersection-Over-Union (IOU) between the predicted box and the ground truth box. Each grid cell also predicts conditional class probabilities, $P(Class_{i} | Object)$, for each of the 18 classes in our task. These probabilities are conditioned on whether the grid cell contains an object. At test time, the conditional class probabilities and the individual box confidence predictions are multiplied as shown in Equation \ref{probeq}, 

\begin{dmath}
P(Class_{i} | Object) \times P(Object) \times IOU^{truth}_{pred}\\
= P(Class_{i}) \times IOU^{truth}_{pred}
\label{probeq}
\end{dmath}

which provides the class-specific confidence scores for each of the box predictions. These scores encompass both the probability of a class being in the box and how accurately the predicted box encloses the object. Our YOLOv3 model predicts bounding boxes at three different scales so, the values for $S$ are $19$, $38$ and $76$. The feature extractor of our model is based on the Darknet-53 architecture. It consists of 53 convolutional layers with batch normalization, residual and shortcut connections as regularization. Fig. \ref{proposedmodel} shows the overall architecture of our model. \par
The YOLOv3 algorithm utilizes pre-defined bounding boxes known as anchor boxes. The height and width of these boxes are defined to capture the aspect ratio and scale of the particular object classes that we wish to detect. During detection, the predefined anchor boxes are convolved across the image. The network predicts the probability and other attributes, such as background, intersection over union (IoU) and offsets for each anchor box. These predictions are used to fine-tune each individual anchor box. We have used nine anchor boxes for our model. To find these anchor boxes, we used k-means clustering on our dataset with k = 9 and using the distance metric given in Equation \ref{distanceeq}.\par
\begin{equation}
    d(box,centroid) = 1 \, \mbox{--} \,  IOU(box,centroid) 
    \label{distanceeq}
\end{equation} \par
The anchor boxes, given as (width, height), we obtained are: $(14, 17)$, $(23, 31)$, $(34, 56)$, $(68, 70)$, $(42,118)$, $(117,111)$, $(105,185)$, $(170,151)$ and $(219,218)$. 

\subsection{Dataset for Training}
In order to train the CNN-based YOLO architecture, we needed a Bengali handwritten digits dataset which contains the images of the digits along with their bounding box annotations. To work on real images, we required variations in their background. The previously created datasets for Bengali handwritten digits recognition did not have any bounding box annotations as they were only intended for classification, not localization. So, we have engineered a new dataset ``Hishab" which contains 40000 images with their bounding box annotations in separate text files. The images in the dataset belong to 18 different classes with each class having nearly 2180 images. The 18 classes are Bengali digits (0-9), operators (addition, subtraction, multiplication  and division), opening and closing brackets, equals sign and decimal point sign.  Also there are 800 images which contain objects belonging to multiple classes. From the dataset, we have used 32000 images for training the model and remaining 8000 as validation set to evaluate the performance of the model. \par
At first, we created a primary dataset and then, used several augmentation techniques on the images to increase variation in the dataset. The following augmentations were applied to the images for regularizing the dataset:
\begin{itemize}
\item{Scaling independently on X axis and Y axis}
\item{Shearing  from -16 degrees to +16 degrees}
\item{Elastic Transformation, which is moving every pixel around in a close local vicinity}
\item{Piecewise Affine, which is placing some points in the image and moving them around with their surrounding areas}
\item{Perspective Transform, which is transforming the digits to view from different angles}
\item{Adding a constant value to one of the color channels}
\item{Swapping the color channels}
\item{Additive Gaussian Noise}
\item{Sharpening}
\item{Blurring} 
\item{Emboss}
\item{Invert}
\end{itemize} \par
\begin{figure}[ht]
    \centering
    \includegraphics[width=\linewidth]{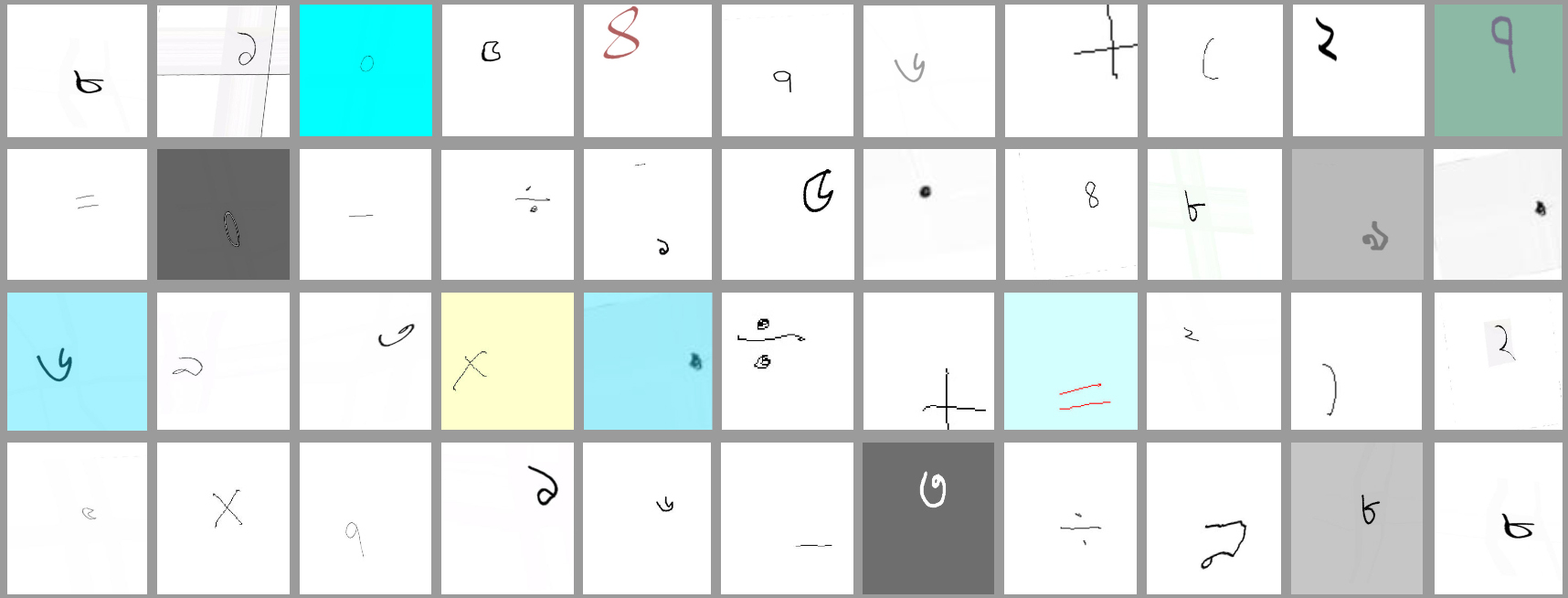}
    \includegraphics[width=\linewidth]{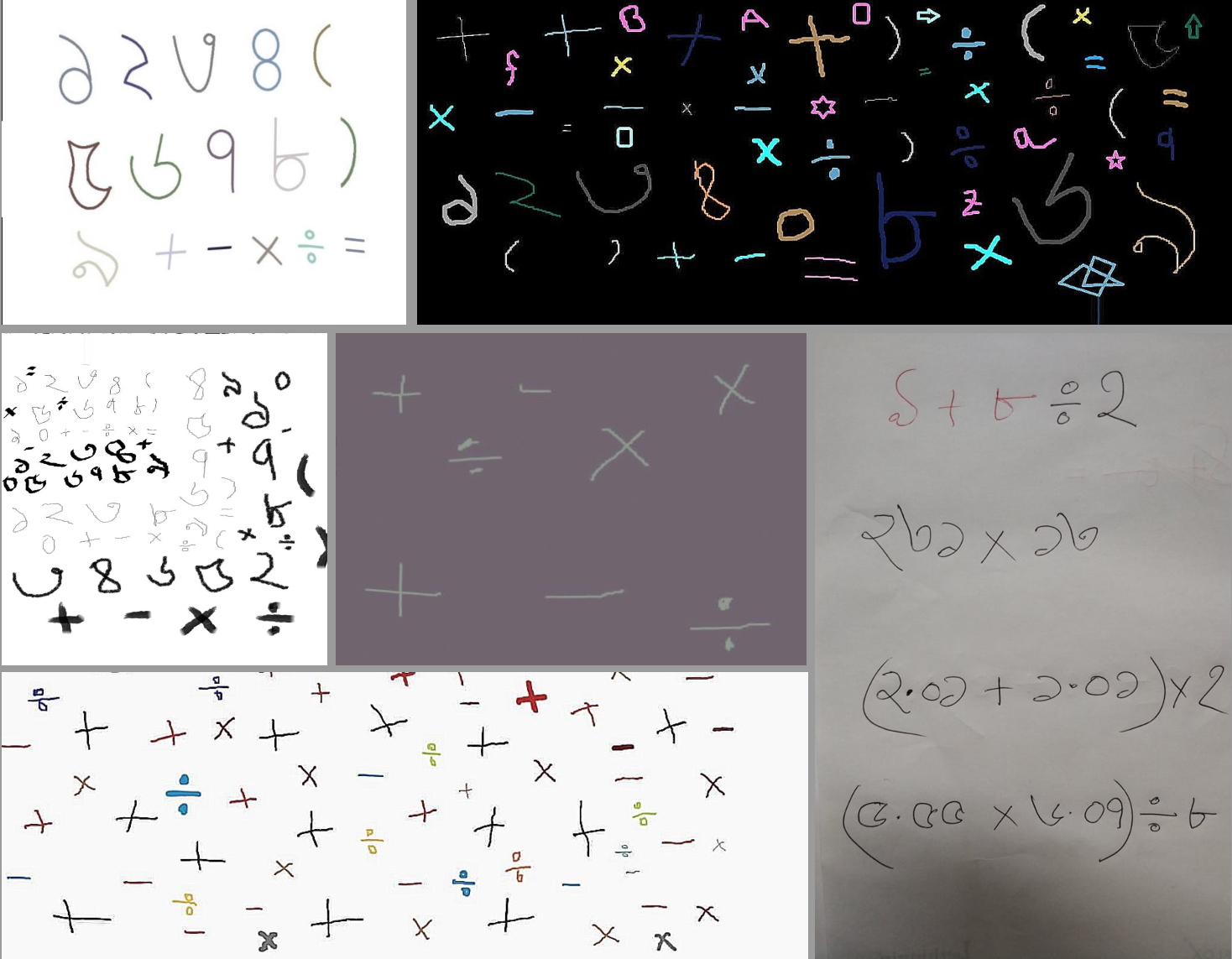}
    \caption{A subset of our dataset}
    \label{subset}
\end{figure}

The ``Hishab" dataset is prepared for object detection and it can aid the researchers to further work on Bengali handwritten digits recognition. Fig. \ref{subset} shows a subset of our dataset.\par
We have also utilized the publicly available NumtaDB \cite{numtadb} and CMATERdb \cite{Das:2012:GAB:2161007.2161320} datasets of handwritten Bengali numerals to assess the efficacy of our feature extractor backbone CNN. From these datasets, we have used 80\% of the images for training the model and the rest as a test set to evaluate the performance of the model. NumtaDB is a dataset containing more than 85000 images of handwritten Bengali numerals accumulated from six different datasets: BHDDB, B101DB, OngkoDB, DUISRT, BanglaLekha Isolated and UIUDB. Being a diverse and unbiased dataset, NumtaDB is ideal for evaluating our backbone CNN. CMATERdb is a balanced dataset of total 6000 handwritten Bangla numerals with 600 images per digit.

\subsection{Training}
\begin{figure}[ht]
    \centering
    \includegraphics[width=\linewidth]{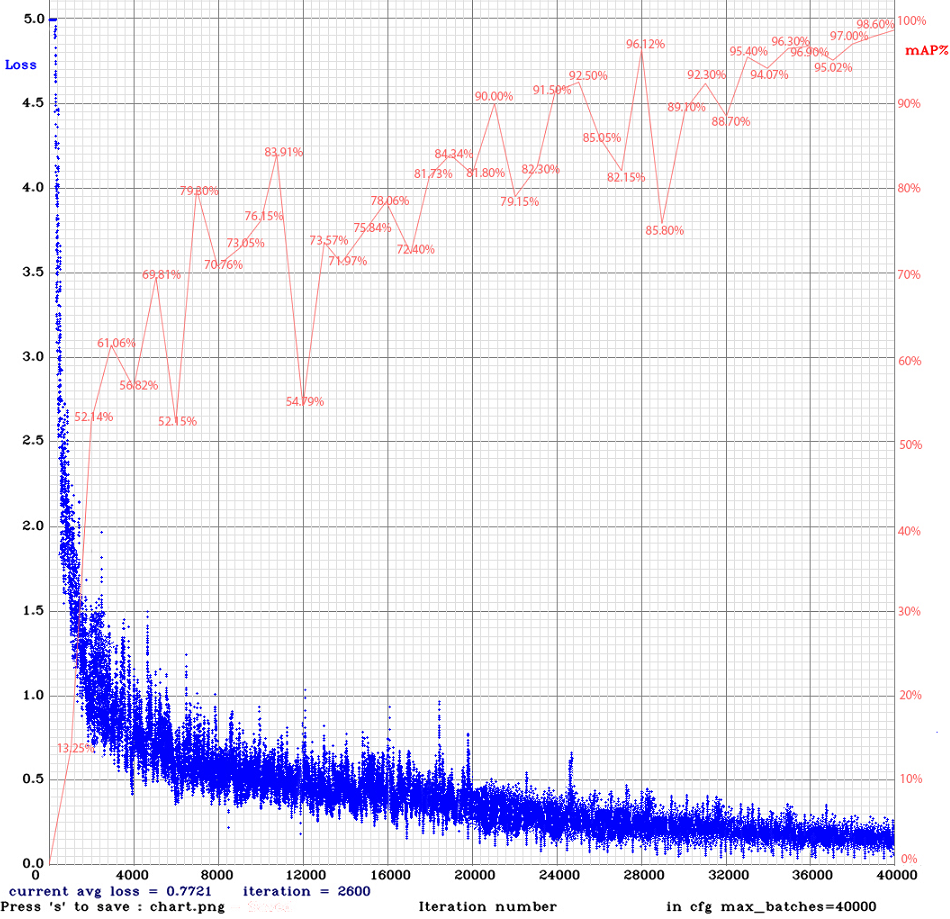}
    \caption{Plot of Average Training Loss and Overall Validation mAP against number of iterations}
    \label{trainingGraph}
\end{figure}

During training, we used batch gradient descent with a batch size of 64 images. The model was trained for 40000 iterations where each iteration is a batch gradient descent step using 64 images. This means we have trained for 64 epochs as our dataset consists of 40000 images. We utilized an SGD optimizer with an initial learning rate of 0.001, momentum of 0.9 and step-wise decays of 0.1 at iteration number 32000 and 36000. The weights for our network were saved after each 1000 iteration. Fig. \ref{trainingGraph} shows the plot of average training loss and validation mean average precision against iteration number for our training. \par
For evaluating our feature extractor backbone CNN, we added a softmax layer over it and trained it using batch gradient descent, with a batch size of 128, on: NumtaDB, BanglaLekha Isolated and CMATERdb. For NumtaDB, we trained for 8 epoch using an SGD optimizer with an initial learning rate of 0.1, momentum of 0.9 and polynomial decay of 0.0005 with a power of 4. For CMATERdb, we trained for 60 epochs using the same optimizer settings as we did for training NumtaDB.

\subsection{Performing Mathematical Operations}
The YOLO model was trained to recognize and localize digits, operators, brackets and decimal points. But this work is not confined to just recognizing the digits and localizing them. To perform the mathematical operations, our system had to understand the numbers from series of digits and locate the position of the operators within the numbers. For the task, the distance from the left vertical axis to the center of each of the object was calculated. This in turn allows our system to pinpoint the positions of each of the digits in the number as well as the operators, brackets and decimal points in the image. With that information, our system is able to figure out the numbers and perform the mathematical operations.  
The model is able to identify and evaluate multiple mathematical expressions given to it in the same image as shown in Fig. \ref{equationFig}. For the task, the upper and lower margins of the first digit from the left vertical axes was calculated. All the objects whose center fall within these two margins are considered to be in the same expression. In this way, the model is able to differentiate between separate mathematical expressions. The strategy to separate out different mathematical expressions is summarized in Algorithm \ref{sepmathexp}.
\begin{algorithm}
\caption{Separation of Mathematical Expressions}\label{sepmathexp}
\begin{algorithmic}
\Require{List containing all detections, $\mbox{I}_{i=1;j=1}^{N\mbox{ }\mbox{ }\hspace{-1pt};5}$ where $i$ selects among the $N$ detections in the image and $j$ selects among $(x_{center}, y_{center}, w, h)$.} \newline \Comment{$(x_{center}, y_{center})$: center of the bounding box} \newline \Comment{$(w, h)$: width and height of the bounding box}
\Ensure{List with separate mathematical expressions, R}
\State $\mbox{R} \gets \emptyset$
\State Sort I according to $x_{center}$, $\mbox{I}_{i;j=2}$ 
\While {$N\neq0$}
    \State $Y_{min}\gets\mbox{I}_{i=1;j=2}\mbox{ -- }0.5\times\mbox{I}_{i=1;j=4}$
    \State $Y_{max}\gets\mbox{I}_{i=1;j=2}+0.5\times\mbox{I}_{i=1;j=4}$
    \State $\mbox{Eq}\gets\emptyset$
    \For{$k\gets N$ to $1$}
        \If{$Y_{min}\leq\mbox{I}_{i=k;j=4}\leq Y_{max}$}
            \State $\mbox{Eq}\gets Join(\mbox{I}_{i=k}, \mbox{Eq})$
            \State $\mbox{I}\gets Join(\mbox{I}_{i=1}^{k-1}, \mbox{I}_{i=k+1}^{N})$
            \State $N\gets N\mbox{ -- }1$
        \EndIf
    \EndFor
    \State $\mbox{R}\gets Join(\mbox{R}, \mbox{Eq})$
\EndWhile
\end{algorithmic}
\algcomment{$Join(\mbox{A}, \mbox{B})$ returns a list containing the elements of list A followed by the elements of list B.}
\end{algorithm}

\section{Results and Evaluation}
The model was trained until the average training loss was less than 0.06. The average validation mAP over all classes after the completion of training was 98.6\%.  \par
\begin{figure}[!htp]
    \centering
    \begin{subfigure}[t]{0.48\linewidth}
        \includegraphics[width=\textwidth]{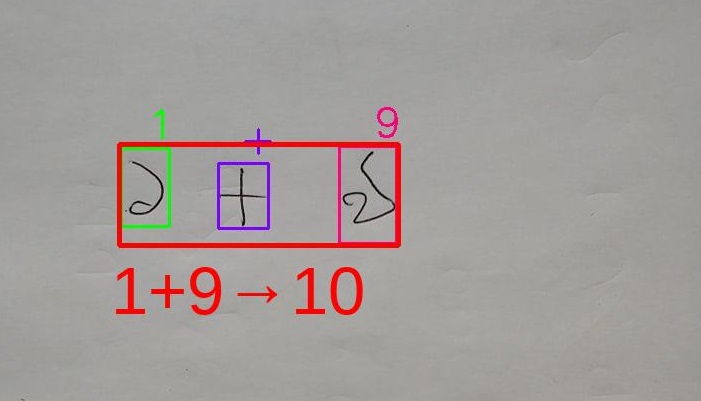}
    \end{subfigure}
    \hspace{-5pt}
    \begin{subfigure}[t]{0.48\linewidth}
        \includegraphics[width=\textwidth]{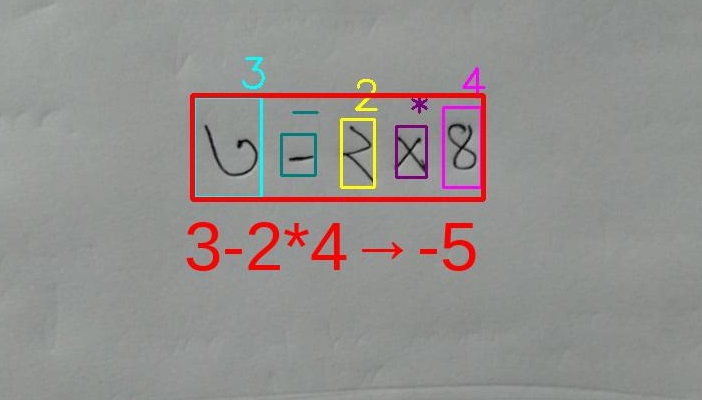}
    \end{subfigure} \par
    \vspace{2pt}
    \begin{subfigure}[t]{0.48\linewidth}
        \includegraphics[width=\textwidth]{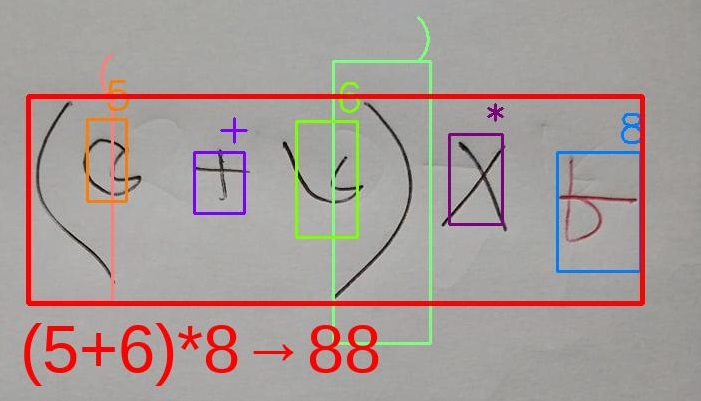}
    \end{subfigure}
    \hspace{-5pt}
    \begin{subfigure}[t]{0.48\linewidth}
        \includegraphics[width=\textwidth]{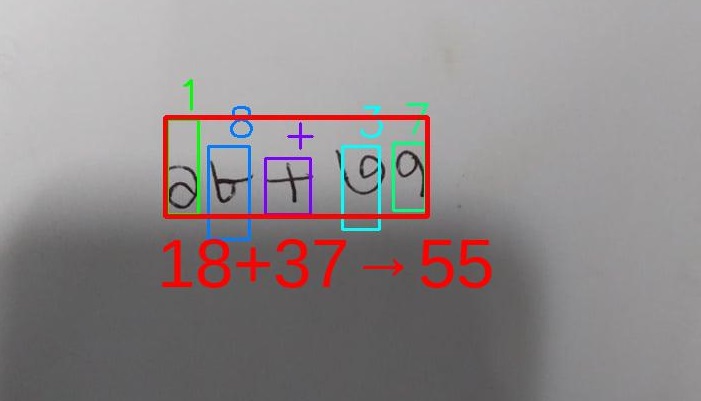}
    \end{subfigure} \par
    \vspace{2pt}
    \begin{subfigure}[t]{0.48\linewidth}
        \includegraphics[width=\textwidth]{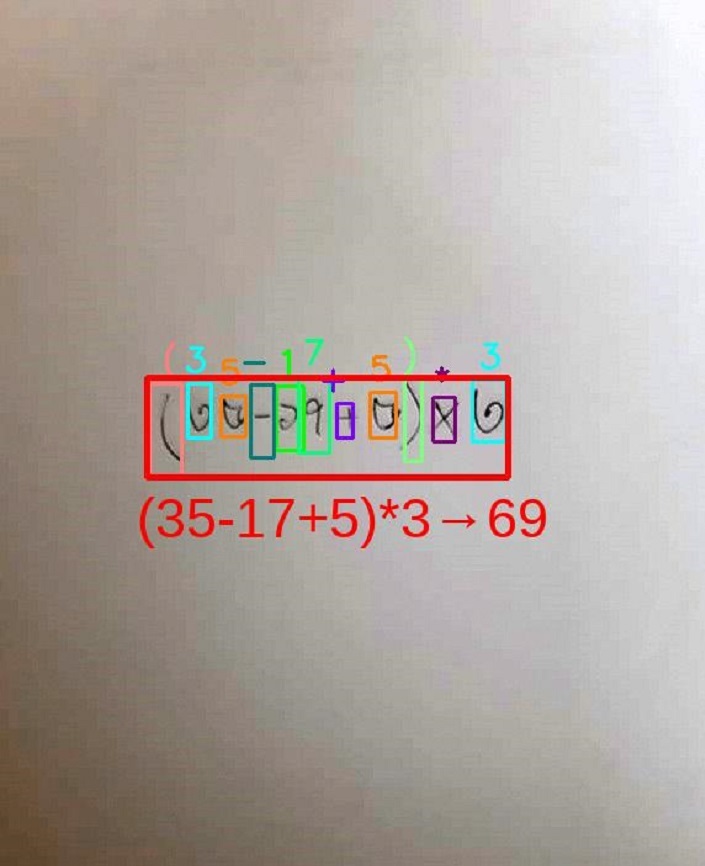}
    \end{subfigure}
    \hspace{-5pt}
    \begin{subfigure}[t]{0.48\linewidth}
        \includegraphics[width=\textwidth]{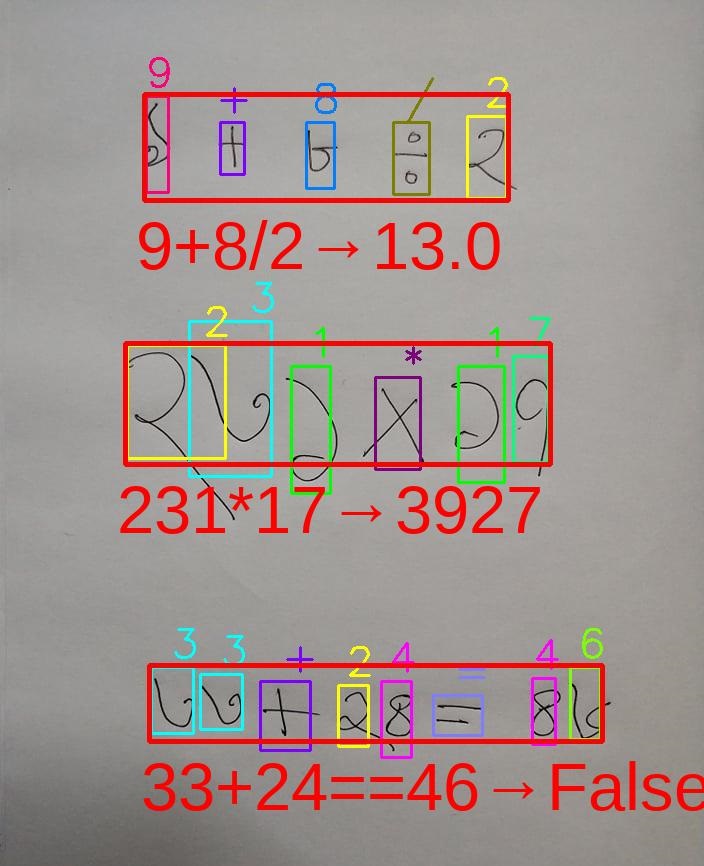}
    \end{subfigure}
    \caption{Different detection outputs for our system}
    \label{equationFig}
\end{figure}

To evaluate our model’s performance on mathematical operations, we have used a test set of 140 images containing mathematical expressions of various complexities; some images contain just two single digit numbers with a single operator while some of the other images contain multiple mathematical expressions with multiple digit decimal numbers and multiple operators and brackets. The performance of the model on this test set is summarized in Table \ref{comptable}. Fig. \ref{equationFig} shows the model's detection outputs on some of the images taken from the test set. \par
To evaluate our choice of the backbone CNN used in our model, we assessed the backbone CNN's performance on two publicly available handwritten Bengali numeral datasets: NumtaDB and CMATERdb. Our backbone CNN achieved a test set accuracy of 99.6252\% on NumtaDB and 99.0833\% on the CMATERdb dataset. \par
\begin{table}[!htb]
\bgroup
\centering
\setlength\tabcolsep{0.019\linewidth}
\begin{tabular}{|l|l|l|l|}
\hline
Category                                                                                                      & Example                                                                                                      & \begin{tabular}[c]{@{}l@{}}Number of\\ Images\end{tabular} & \begin{tabular}[c]{@{}l@{}}Correct\\Predictions\end{tabular} \\ \hline
\begin{tabular}[c]{@{}l@{}}Single digit \\ numbers with \\ single operator\end{tabular}                     & $2 + 3$                                                                                                      & 20                                                          & 20                                                                         \\ \hline
\begin{tabular}[c]{@{}l@{}}Single digit\\ numbers with \\ multiple operators\end{tabular}                  & $3 \, \mbox{--} \, 1 \times 2$                                                                                             & 20                                                          & 20                                                                         \\ \hline
\begin{tabular}[c]{@{}l@{}}Single digit \\ numbers with \\ multiple operators \\ and brackets\end{tabular} & $(3 + 7 + 5) \div 4$                                                                                         & 20                                                          & 19                                                                         \\ \hline
\begin{tabular}[c]{@{}l@{}}Double digit \\ numbers with \\ single operator\end{tabular}                    & $21\, \mbox{--} \,15$                                                                                                    & 20                                                          & 20                                                                         \\ \hline
\begin{tabular}[c]{@{}l@{}}Double digit \\ numbers with \\ multiple operators \\ and brackets\end{tabular}  & $(42 \, \mbox{--} \, 47) \div 7$                                                                                          & 20                                                          & 19                                                                         \\ \hline
\begin{tabular}[c]{@{}l@{}}Numbers with \\ decimal point\end{tabular}                                       & $(2.54 + 5.55) \times 2$                                                                                     & 20                                                          & 18                                                                         \\ \hline
\begin{tabular}[c]{@{}l@{}}Images with \\ multiple \\ mathematical \\ expressions\end{tabular}               & \begin{tabular}[c]{@{}l@{}}$(2 + 7 \, \mbox{--} \, 5) \times 33.2$\\ $9 + 4 \, \mbox{--} \, 2$\\ $(3 + 4 \, \mbox{--} \, 5) \div (6 + 7)$\end{tabular} & 20                                                          & 16                                                                         \\ \hline
\end{tabular}
\egroup
\caption{Evaluation of our model with different categories of images.}
\label{comptable}
\end{table}

\section{Conclusion}
This research presents a CNN-based object detection model which can detect and localize Bengali handwritten digits and operators, recognize numbers form series of digits using their localization information, and perform mathematical operations on the numbers. Our work, unlike previous research works, treats the task as an object recognition and localization task which enables the model to recognize and calculate multiple mathematical expressions. Another vital contribution of this work is the ``Hishab" dataset which is the first Bengali handwritten digits dataset engineered for object detection containing bounding box annotations. We have also proposed a CNN architecture, utilized as the feature extractor backbone for our model, which outperforms previous classifiers on handwritten Bengali digit recognition.

\bibliographystyle{./bibliography/IEEEtran}
\bibliography{./bibliography/IEEEabrv}

\end{document}